\documentclass[a4paper, 10pt, conference]{IEEEtran}
\IEEEoverridecommandlockouts

\usepackage{cite}
\usepackage{amsmath,amssymb,amsfonts}
\usepackage{algorithmic}
\usepackage{graphicx}
\usepackage{textcomp}
\usepackage{hyperref}
\usepackage{kotex}
\usepackage{multirow}
\usepackage{longtable}
\usepackage{subfigure}
\usepackage{xcolor}
\def\BibTeX{{\rm B\kern-.05em{\sc i\kern-.025em b}\kern-.08em
    T\kern-.1667em\lower.7ex\hbox{E}\kern-.125emX}}
\begin{document}

\title{HTNet: Anchor-free Temporal Action Localization
with Hierarchical Transformers\\
}

\author{\IEEEauthorblockN{Tae-Kyung Kang}
\IEEEauthorblockA{\textit{Dept. Artificial Intelligence} \\
\textit{Korea University}\\
Seoul, South Korea \\
tk\_kang@korea.ac.kr}
\and
\IEEEauthorblockN{Gun-Hee Lee}
\IEEEauthorblockA{\textit{Dept. Computer Engineering} \\
\textit{Korea University}\\
Seoul, South Korea \\
gunhlee@korea.ac.kr}
\and
\IEEEauthorblockN{Seong-Whan Lee}
\IEEEauthorblockA{\textit{Dept. Artificial Intelligence} \\
\textit{Korea University}\\
Seoul, South Korea \\
sw.lee@korea.ac.kr}
}

\maketitle

\begin{abstract}

Temporal action localization (TAL) is a task of identifying a set of actions in a video, which involves localizing the start and end frames and classifying each action instance. Existing methods have addressed this task by using predefined anchor windows or heuristic bottom-up boundary-matching strategies, which are major bottlenecks in inference time. Additionally, the main challenge is the inability to capture long-range actions due to a lack of global contextual information. In this paper, we present a novel anchor-free framework, referred to as HTNet, which predicts a set of $\langle$start time, end time, class$\rangle$ triplets from a video based on a Transformer architecture. After the prediction of coarse boundaries, we refine it through a background feature sampling (BFS) module and hierarchical Transformers, which enables our model to aggregate global contextual information and effectively exploit the inherent semantic relationships in a video. We demonstrate how our method localizes accurate action instances and achieves state-of-the-art performance on two TAL benchmark datasets: THUMOS14 and ActivityNet 1.3.

\end{abstract}
\section{Introduction}

Temporal action localization (TAL) is a task that localizes action instances in a video by predicting the start and end times, as well as the class. Recently, videos have been untrimmed and long, unlike trimmed and short video clips; therefore, the TAL task can be used for many applications, including video analysis, summary, and human interaction~\cite{PR1,PR3,PR9}. However, this is challenging compared to an action recognition task~\cite{PR4,PR7, PR10} that simply predicts the action class of short videos. This is because the length of action instances in videos is diverse; therefore, it is difficult to estimate each action instance's start and end times. To solve this problem, existing methods use predefined anchors~\cite{GTAN,TAL,PR2} or estimate the actionness of each frame~\cite{BSN,BMN,BSN++,smc1}. 

Anchor-based methods generate action proposals based on a dense box placement. However, the lengths of action instances can vary from several seconds to several minutes; thus, it is almost impossible to cover all ground-truth instances under reasonable computation consumption. Therefore, these approaches are sensitive to parameters such as the size or number of anchors, and their computational costs are high.
A few anchor-free approaches~\cite{AFSD,smc2} have been suggested to overcome these limitations by generating only one proposal on the temporal locations. However, their performances are unsatisfactory for use in real applications. These methods are not dependent on predefined anchors and the actionness of each frame. Hence, this anchor-free approach emphasizes the estimation of meaningful features and semantic relationships in a video. We argue that previous works on TAL are limited in capturing features, including both local boundary information and long-range temporal information. This lack of capturing both information can cause the inability to predict long-range actions, negatively affecting model performance.

We propose a coarse-to-fine anchor-free architecture based on transformers to predict a set of TAL triplets, enabling the model to overcome the limitations of previous studies. First, a background feature sampling (BFS) module allows refinement of the features to include local boundary information and long-range temporal information. Coarse temporal features are integrated with boundary-attentive features and long-range temporal features from hierarchical temporal sampling. Second, we designed hierarchical transformers with multi-level temporal features, making the model exploit the inherent semantic relationships in a video. These allow the model to consider the temporal semantics of the video further.

We evaluate our model using two TAL benchmarks, THUMOS14 and ActivityNet 1.3. The qualitative and quantitative results show that our method outperforms previous state-of-the-art methods on the two datasets. The contributions of this work can be summarized as follows:

\begin{itemize}

\item To address the inability to capture long-range actions, we propose a hierarchical transformer-based TAL set prediction approach called HTNet, which enables the refinement of coarse features to include local boundary and long-range temporal information.

\vspace{.3cm}

\item We propose a background feature sampling (BFS) module with hierarchical temporal sampling to consider the inherent temporal semantics of a video.

\vspace{.3cm}

\item HTNet achieves state-of-the-art performance on primary benchmark datasets in the TAL task: THUMOS-14 and ActivityNet1.3.

\end{itemize}

\begin{figure*}[t!]
    \centering
    \includegraphics[width=0.8\linewidth]{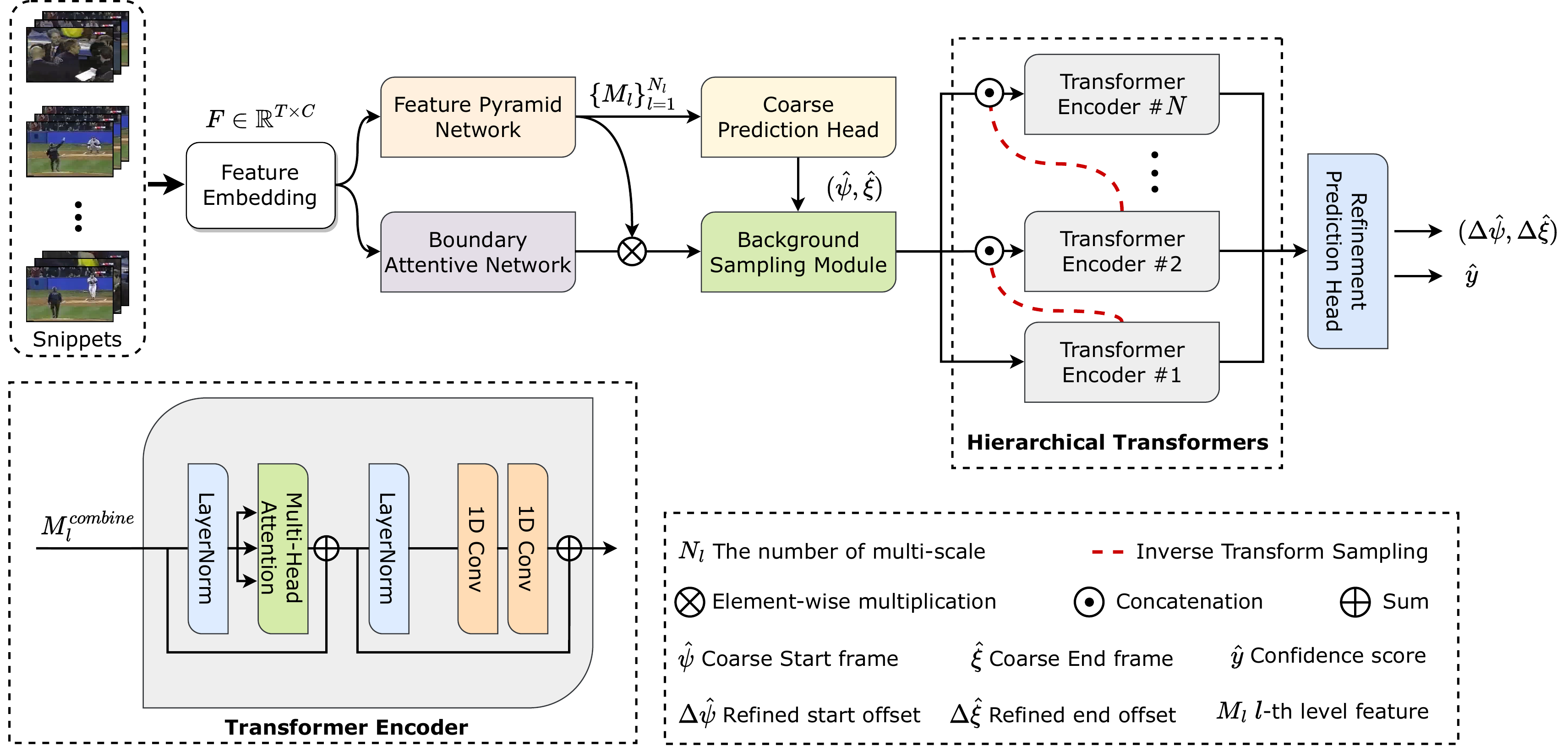}
   
    \caption{The overall framework of our HTNet. Given a video, we extract multi-scale features and then predict coarse boundaries. Next, we obtain background context through a Background Feature Sampling (BFS) module based on the coarse boundaries. Finally, we predict a set of fine-grained boundaries and classes using encoded features by Hierarchical Transformers.}
    \label{method}
\end{figure*}

\section{Related Work}

\subsection{Temporal Action Localization}
Temporal action localization aims to detect actions in untrimmed videos, regress boundaries, and classify classes of actions. Existing temporal action localization approaches can be divided into one-stage~\cite{SSN,GTAN,PBRNet,AFSD} and two-stage strategies~\cite{MUSES,DCAN,ContextLoc,RTD-Net}, similar to object detection~\cite{PR5}. One-stage approach directly predicts both action boundaries and classes. On the other hand, the two-stage approach first generates action proposals. It then classifies the action classes of the proposals. Most TAL methods adopt a two-stage bottom-up approach called actionness-guided. Actionness-guided methods aim to generate action proposals by evaluating the actionness that denotes the probability of action for each frame. For example, BSN~\cite{BSN} predicts the actionness of each temporal location and then aggregates the locations with high start and end probabilities to generate the proposals. Similarly, BMN~\cite{BMN} generates proposals using a Boundary-Matching confidence map, allowing for better proposals. However, these methods have two main limitations: (1) These methods consider all possible combinations of temporal locations. (2) They have to combine with other video-level classification networks. In contrast, our method directly predicts boundaries for each temporal location through a single network.

\subsection{Transformer}
Current transformer architectures outperform in various fields such as natural language processing (NLP) and computer vision. Generally, transformer aims to relieve the problem of long-range dependency modeling in sequential tasks by utilizing a self-attention mechanism. For the first time in the computer vision field, ViT~\cite{ViT} adopts Transformer architecture that splits 2D images into multiple patches and then conducts self-attention. In the TAL task, RTD-Net~\cite{RTD-Net} uses a transformer decoder to obtain sparse proposals without post-processing. Likewise, the self-attention mechanism is used to exploit large-scale or long-range contexts. Inspired by these advancements, we design a transformer-based architecture to model the inherent semantic relationships between each temporal location.

\section{Method}

\subsection{Problem Definition}
In the temporal action localization task, input is an untrimmed video \(X=\{ x_i \in \mathbb{R}^{C \times H \times W} \}_{i=1}^{T} \) with $T$ frames, and output is a set of the temporal action boundaries and classes \(P=\{\hat{\psi}_{j},\hat{\xi}_{j},\hat{y}_{j}\}_{j=1}^{N_{P}}\), where \(\hat{\psi}_{j}, \hat{\xi}_{j}, \hat{y}_{j},\) and \(N_{P}\) denote the start time, end time, action class, and the number of actions in the video, respectively.

\subsection{Feature Extraction}
We extract a feature using a pre-trained I3D~\cite{I3D} model given the untrimmed video \(X\). The extracted feature \(F\in \mathbb{R}^{T\times C}\) is divided into \(N\) multi-scale features $\{M_l\in \mathbb{R}^{T_l \times C}\}_{l=1}^{N}$ by \(1D\) convolution operations to have various temporal dimensions like a pyramid structure. The higher the feature level, the smaller the time dimension is as follows:
\begin{equation}
    T_l=\frac{T_1}{2^{l-1}},
\end{equation}
where $l\in \{1,2,\dots,N\}$.
For convenience, we will explain using the $l$-th level feature $M_{l}$.

\subsection{Coarse Prediction}
We use a basic anchor-free prediction module to obtain coarse temporal boundaries of action instances. Given the multi-scale features \(\{ M_l \in \mathbb{R}^{T_l \times C} \}_{l=1}^{N}\), simple regression heads predict coarse start and end boundary distances \((d_i^s, d_i^e)\) for each location \(i \in \{1,2,\dots,T_l\}\). Then, we can obtain \(\hat{\psi}_{i}=i-d_{i}^{s}*2^{l}\) and \(\hat{\xi}_i=i+d_{i}^{e}*2^{l}\), which denote the start and end frames for the \(i\)-th time step in the \(l\)-th level. In the $l$-th level, we can obtain \(T_{l}\) proposals. The anchor-free approach allows the model to predict action instances without predefined anchors, which generates fewer proposals but more accurate temporal action boundaries.

\subsection{Boundary-attentive network}
We adopt a boundary-attentive network to make the multi-scale features keep boundary information. Boundary-attentive network generates the feature representing the start and end location probabilities as follows:
\begin{equation}
\begin{aligned}
&B^{s}=\sigma(\mathrm{LN}(\mathrm{Conv1d}(M_1))) \quad \in \mathbb{R}^{T \times C}, \\
&B^{e}=\sigma(\mathrm{LN}(\mathrm{Conv1d}(M_1))) \quad \in \mathbb{R}^{T \times C},
\end{aligned}
\end{equation}
where \(\sigma\), and \(\mathrm{LN}\) denote ReLU activation function, and layer normalization, respectively. These are processed from the first level feature $M_1$; thus, we conduct the max-pooling operation to fit the temporal dimension \(T_l\) of each scale level \(l\). The boundary-attentive feature explicitly enhances the features to keep sharp boundary information, as previously discussed in~\cite{RTD-Net}.

\subsection{Background Feature Sampling}
To address the lack of long-region action proposal generation, we propose a background feature sampling (BFS) module aggregating global background context. We design BFS based on the intuition that background regions are complementary information to each boundary proposal. BFS has three inputs: multi-scale features, boundary-attentive features, and coarse boundaries. Finally, we can obtain combined features aggregated with global context and enhanced by boundary-attentive features.

\begin{figure}
\centering
\includegraphics[width=1.0\linewidth]{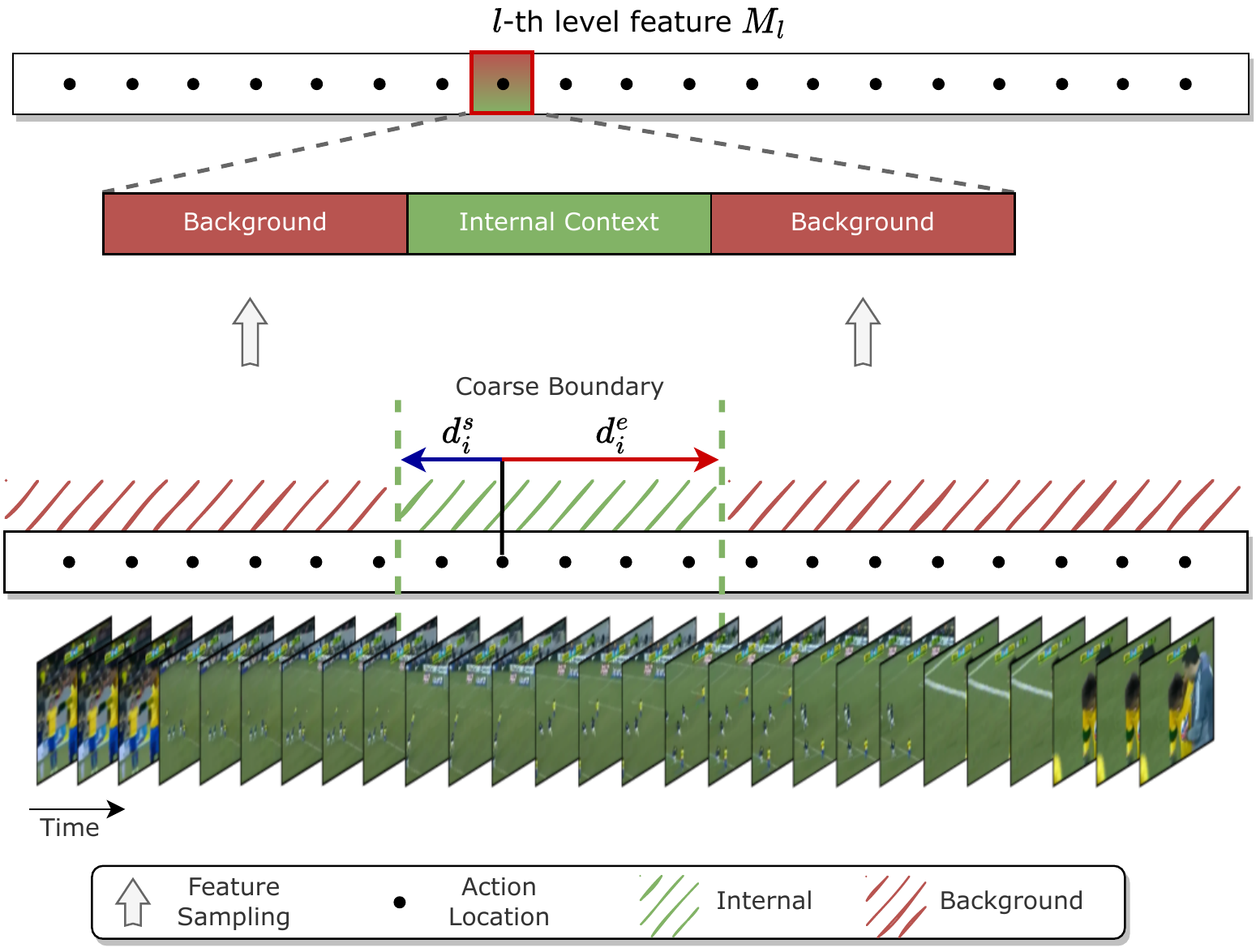}
\caption{The architecture of Background Feature Sampling (BFS) module. We sample background features $M_l^{BG}$ based on the coarse boundaries, and conduct channel-wise concatenation with $M_l$.}
\label{BFS}
\end{figure}
\subsubsection{\textbf{Sampling based on feature importance}}
We propose a novel feature sampling method to aggregate the context of the background region. First, we divide temporal regions on the $k$-th location of the $l$-th level feature into two groups based on the coarse boundaries: left-background, right-background, as shown in Fig.~\ref{BFS}. We then sample the significant features using max pooling operation for each group at different rates $\delta$. Specifically, we conduct an ablation study in Tab.~\ref{tab3} to find an optimal sampling range (left-background: $[\delta \cdot \hat{\psi},\hat{\psi}]$ or right-background: $[\hat{\xi}, \delta \cdot (T_l-\hat{\xi})]$) and set $\delta$ to 0.7. The sampling method works as follows:
\begin{equation}
\begin{aligned}
&M_l^{left}=\bigcup_{i=0}^{T_l}{\max_{j \in B_{left}^{i}}{M_l(j)}}, \\
&M_l^{right}=\bigcup_{i=0}^{T_l}{\max_{j \in B_{right}^{i}}{M_l(j)}},
\end{aligned}
\end{equation}
where $B_{left}^i$ and $B_{right}^i$ denote left-background and right-background ranges for $i$ location. After sampling for each location, we obtain a background feature \(M_l^{BG}\) stacked with the maximum values within the region:
\begin{equation}
M_l^{BG}=M_l^{left} \odot M_l^{right} \quad \in \mathbb{R}^{T_{l} \times 2C},
\end{equation}
where \(\odot\) denotes channel-wise concatenation.

\subsubsection{\textbf{Feature refining}}
Finally, we can obtain combined features utilizing multi-scale, start-end, and context-aware features. We first conduct an element-wise sum with the start $B^s$ and end $B^e$ features to generate boundary-attentive features. Then, we conduct element-wise multiplication with multi-scale feature and boundary-attentive feature so that the feature has attention around boundaries. Here, we add the aggregated global context $M_{l}^{BG}$ to $M_{l}$ by conducting channel-wise concatenation; we then use temporal convolution to reduce channels and build a combined feature $\textbf{H}_l$:
\begin{equation}
\begin{aligned}
&\textbf{H}_{l}=\mathrm{Conv}(M_l \odot M_{l}^{BG}) \quad \in \mathbb{R}^{T_l \times C},
\end{aligned}
\end{equation}
where $\odot$ denotes channel-wise concatenation. We utilize this combined feature as input of hierarchical transformer to model the inherent semantic relationships between action instances.

\subsection{Fine-grained Prediction with Hierarchical Transformers}
We propose hierarchical transformers to refine features to include inherent semantic relationships in a video for fine-grained regression. By designing the transformer hierarchically, our network further considers temporal semantics. We only adopt transformer encoders because the role of hierarchical transformers is simply modeling the semantic relationships. We describe each transformer structure as follows.

\subsubsection{\textbf{Hierarchical structure}}

Our network has $N$ transformer encoders equal to the number of multi-scale to preserve temporal semantics for each multi-scales. In addition, we note that the multi-scale features lose detailed temporal information. Therefore, we sample the contextual information in previous level features and concatenate it to the following features as shown in Fig.~\ref{method}. Here, we adopt the inverse transform sampling algorithm considering the whole context of the features. Specifically, we produce the probability density function from $\textbf{H}_{l}$ by conducting a channel-wise mean. Then, we sample an index set along the temporal dimension from this distribution using the inverse transform sampling and obtain the sampled feature by taking the values of the corresponding index. Finally, these features are encoded as proposal features \(\textbf{Z}_l\) by transformer encoders for each level.

\subsubsection{\textbf{Self-attention}}
The input $\textbf{H}_l \in \mathbb{R}^{T_l\times C}$ is projected using $\mathrm{\textbf{W}}_Q \in \mathbb{R}^{C \times C_q}$, $\mathrm{\textbf{W}}_K \in \mathbb{R}^{C \times C_k}$, and $\mathrm{\textbf{W}}_V \in \mathbb{R}^{C \times C_v}$ to extract feature representations query $\textbf{Q}$, key $\textbf{K}$, and value $\textbf{V}$, respectively.
The outputs $\textbf{Q}$, $\textbf{K}$, $\textbf{V}$ are computed as:
\begin{equation}
\begin{aligned}
&\textbf{Q}=\mathrm{\textbf{H}}_l\textbf{W}_Q,\\
&\textbf{K}=\mathrm{\textbf{H}}_l\textbf{W}_K,\\
&\textbf{V}=\mathrm{\textbf{H}}_l\textbf{W}_V.
\end{aligned}
\end{equation}
The output of self-attention is given by,
\begin{equation}
\mathrm{\textbf{S}}=\textrm{softmax}(\frac{\mathrm{\textbf{Q}}\mathrm{\textbf{K}}^T}{\sqrt{C_q}})\mathrm{\textbf{V}}.
\end{equation}

\subsection{Refinement Prediction Heads}
Refinement prediction heads include class classifier and boundary regressor, which consists of $1D$ convolutions as follows:
\begin{equation}
\begin{aligned}
&\hat{y}_l=\mathrm{Classifier}(\sigma(\mathrm{LN}(\mathrm{Conv}(\textbf{Z}_l)))) & \in \mathbb{R}^{T_l \times N_c}, \\
&\hat{\Psi}_l=\mathrm{Regressor}(\sigma(\mathrm{LN}(\mathrm{Conv}(\textbf{Z}_{l})))) \times \omega & \in \mathbb{R}^{T_l \times 2},
\end{aligned}
\end{equation}
where $y_l$, $\Psi_l$, $N_c$, and $\omega$ denote $l$-th level confidence scores, refined distances, the number of classes, and scale factor, respectively. The classifier and regressor produce confidence scores and the refined distances from every moment $T_l$ across all levels on multi-scale, respectively, which have the same design except for the dimension of the final output.

\subsection{Loss function and Inference}

\subsubsection{\textbf{Loss function}}
In this section, we introduce our loss functions. First, the outputs of our method are the coarse distance $(\hat{\psi}_i, \hat{\xi}_i)$, refined distance $(\Delta\hat{\psi}_i,\Delta\hat{\xi}_i)$ from each location $i$, and corresponding confidence score $\hat{y}_i$. Furthermore, we denote target distance as $(\psi_i,\xi_i)$. Our loss function has five terms: (1) $\mathcal{L}_{coarse}$ and (2) $\mathcal{L}_{refine}$ are generalized IoU loss for coarse distance regression. (3) $\mathcal{L}_{cls}$ is a focal loss for multi classification. (4) $\mathcal{L}_{start}$ and (5) $\mathcal{L}_{end}$ are binary cross-entropy loss for boundary-attentive scores. Specifically, we re-scale and take channel-wise mean on the start and end features to obtain \(\tilde{g}^{s}\) and \(\tilde{g}^{e}\), respectively. These two features represent the probability of start and end points. We define the ground truth \(g^{s},g^{e}\) as follows:
\begin{equation}
\begin{aligned}
&g^{s}(i)=\mathbb{I}\left(i \in [\psi-\tau,\psi+\tau]\right), & i=0, 1, \cdots, T_l, \\
&g^{e}(i)=\mathbb{I}\left(i \in [\xi-\tau,\xi+\tau]\right), & i=0, 1, \cdots, T_l,
\end{aligned}
\end{equation}
where \(\mathbb{I}(\cdot)\) and \(\tau\) denote the indicator function and range hyper-parameter, respectively. We set $\tau$ to 5. After that we can calculate the Cross Entropy:
\begin{equation}
\begin{aligned}
&\mathcal{L}_{start}=\mathrm{BCE}\left(\tilde{g}^{s}, g^{s}\right), \\
&\mathcal{L}_{end}=\mathrm{BCE}\left(\tilde{g}^{e}, g^{e}\right),
\end{aligned}
\end{equation}
where \(\mathrm{BCE}\) denotes the binary cross entropy loss. With $\mathcal{L}_{start}$ and $\mathcal{L}_{end}$, we can obtain the feature with high activation at the action area. Finally, we define total loss as follows:
\begin{equation}
\mathcal{L}_{total}=\lambda(\mathcal{L}_{refine} + \mathcal{L}_{coarse}) + \mathcal{L}_{cls} +\mathcal{L}_{start} + \mathcal{L}_{end},
\end{equation}
where $\lambda$ is the hyper-parameter, balancing the classification and regression loss.

\subsubsection{\textbf{Inference}}
In inference, the final outputs $(\tilde{\psi}_i,\tilde{\xi}_i,\tilde{y}_i)$ of our method are built with refined distance $(\Delta\hat{\psi}_i,\Delta\hat{\xi}_i)$, and $\hat{y}_i$ as follows:
\begin{equation}
\begin{aligned}
&\tilde{\psi}_i=i-(\Delta\hat{\psi}_i \times 2^{l}), \\
&\tilde{\xi}_i=i-(\Delta\hat{\xi}_i \times 2^{l}), \\
&\tilde{y}_i=\mathrm{sigmoid}(\hat{y}_i), \\
\end{aligned}
\end{equation}
where $\tilde{\psi}_i$, $,\tilde{\xi}_i$, $\tilde{y}_i$, and $l$ denote the final start frame, end frame, confidence score of temporal location $i$, and the level of features, respectively. Then, we suppress redundant proposals using Soft-NMS.

\begin{figure}[t!]
    \centering
    \includegraphics[width=1.0\linewidth]{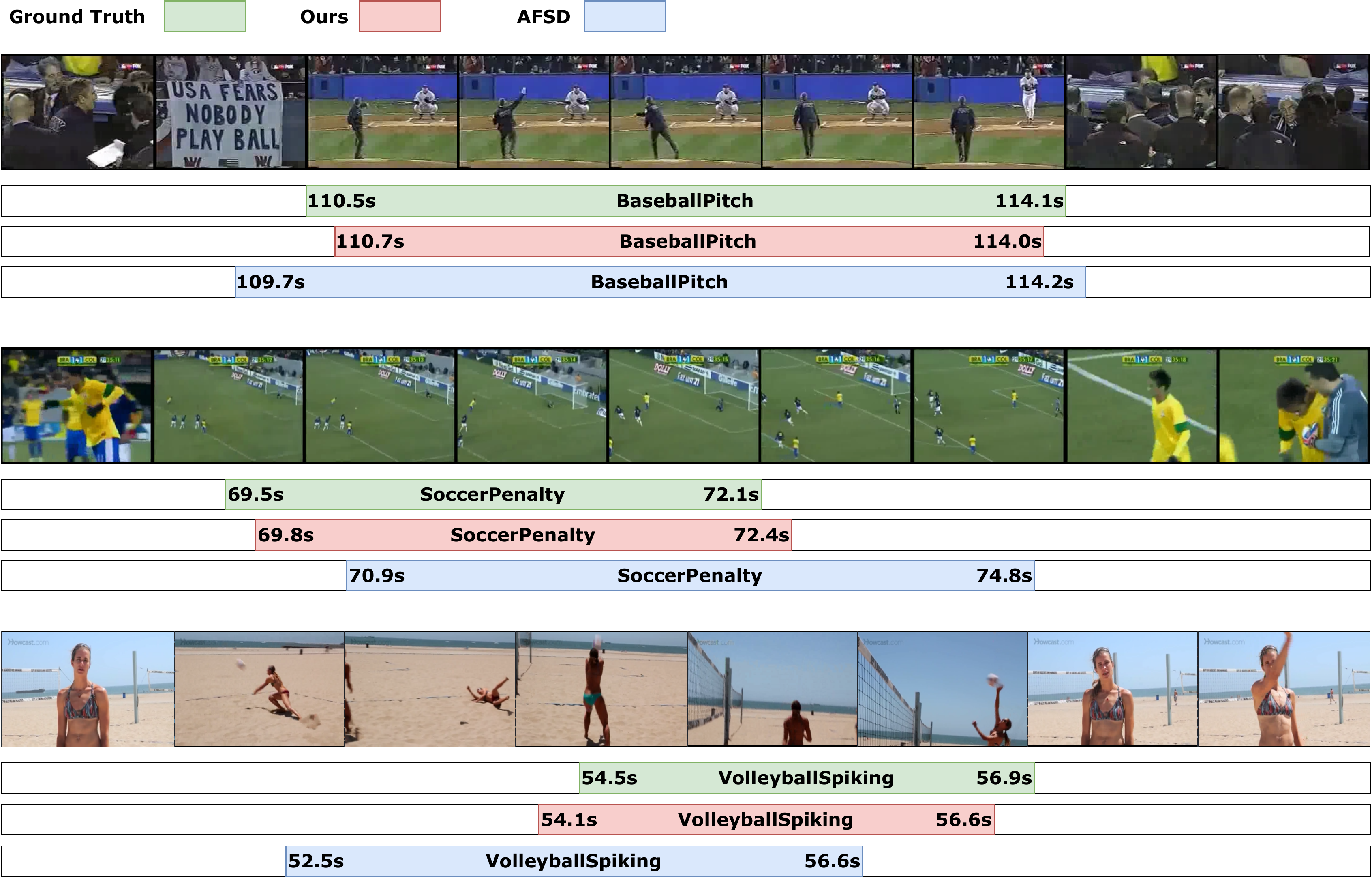}
   
    \caption{Qualitative comparison of the results from our method and the baseline AFSD~\cite{AFSD}. We conduct experiments on three videos; the first video is relatively easy to detect the action because it is visually explicit. In contrast, the second and third videos have similar actions, which are not visually explicit, so tricky to detect the action.}
    \label{qualitative}
\end{figure}
\begin{table*}[t!]
\centering
\caption{Comparison with other state-of-the-art methods on THUMOS-14 and ActivityNet1.3 in terms of mAP (\%).}
\label{tab1}
\resizebox{\textwidth}{!}{%
\begin{tabular}{c|c|c|cccccc|cccc}
\hline
\multirow{2}{*}{Type} &
  \multirow{2}{*}{Method} &
  \multirow{2}{*}{Feature} &
  \multicolumn{6}{c|}{THUMOS-14~\cite{D1}} &
  \multicolumn{4}{c}{ActivityNet1.3~\cite{D2}} \\ \cline{4-13} 
                            &                                            &     & 0.3  & 0.4  & 0.5  & 0.6  & 0.7  & Avg.  & 0.5   & 0.75  & 0.95 & Avg.  \\ \hline
\multirow{10}{*}{Two stage} 
                            & TAL~\cite{TAL}       & I3D & 53.2 & 48.5 & 42.8 & 33.8 & 20.8 & 39.8  & 38.2  & 18.3  & 1.3  & 20.2  \\
                            & BSN~\cite{BSN}       & TS  & 54.5 & 45.0 & 36.9 & 28.4 & 20.0 & 36.8  & 46.5  & 30.0  & 8.0  & 30.0  \\
                            & BMN~\cite{BMN}       & TS  & 56.0 & 47.4 & 38.8 & 29.7 & 20.5 & 38.5  & 50.1  & 34.8  & 8.3  & 33.9  \\
                            & G-TAD~\cite{G-TAD}   & TS  & 54.5 & 47.6 & 40.2 & 30.8 & 23.4 & 39.3  & 50.4  & 34.6  & 9.0  & 34.1  \\
                            & BSN++~\cite{BSN++}   & TS  & 59.9 & 49.5 & 41.3 & 31.9 & 22.8 & 41.0  & 51.2  & 35.7  & 8.3  & 34.8  \\
 &
  RTD-Net~\cite{RTD-Net} &
  TSN,I3D &
  68.3 &
  62.3 &
  51.9 &
  38.8 &
  23.7 &
  49.0 &
  47.2 &
  30.6 &
  8.6 &
  30.8 \\
 &
  ContextLoc~\cite{ContextLoc} &
  I3D &
  68.3 &
  63.8 &
  54.3 &
  41.8 &
  26.2 &
  50.8 &
  \textbf{56.0} &
  35.2 &
  3.6 &
  34.2 \\
 &
  DCAN~\cite{DCAN} &
  TS &
  68.2 &
  62.7 &
  54.1 &
  43.9 &
  32.6 &
  52.3 &
  51.7 &
  35.9 &
  \textbf{9.4} &
  35.3 \\
                            & MUSES~\cite{MUSES}   & I3D & 68.9 & 64.0 & 56.9 & 46.3 & 31.0 & 53.4  & 50.0 & 34.9 & 6.5 & 33.9 \\ \hline
\multirow{5}{*}{One stage}  
& SSN~\cite{SSN}       & TS  & 51.0 & 41.0 & 29.8 & -    & -    & -     & 43.2  & 28.7  & 5.6  & 23.8  \\
                            & GTAN~\cite{GTAN}     & P3D & 57.8 & 47.2 & 38.8 & -    & -    & -     & 52.6  & 34.1  & 8.9  & 34.3  \\
                            & PBRNet~\cite{PBRNet} & I3D & 58.5 & 54.6 & 51.3 & 41.8 & 29.5 & 47.14 & 53.9  & 34.9  & 8.9  & 35.0  \\
                            & AFSD~\cite{AFSD}     & I3D & 67.3 & 62.4 & 55.5 & 43.7 & 31.1 & 52.0  & 52.4  & 35.3  & 6.5  & 34.4  \\ \cline{2-13} 
 &
  \textbf{Ours} &
  I3D &
  \textbf{71.2} &
  \textbf{67.2} &
  \textbf{61.5} &
  \textbf{51.0} &
  \textbf{39.3} &
  \textbf{58.0} &
  53.9 &
  \textbf{36.9} &
  8.1 &
  \textbf{36.0}\\
  \hline
\end{tabular}%
}
\end{table*}

\setlength{\tabcolsep}{2.5pt}
\begin{table}[t]
\centering
\caption{Ablation study on the effect of BFS and Hierarchical Transformer in terms of mAP(\%).}
\label{tab2}
{\footnotesize
\begin{tabular}{c|c|cccc|cccc}
\hline
\multirow{2}{*}{\begin{tabular}[c]{@{}c@{}}Backgorund\\ Sampling\end{tabular}} &
  \multirow{2}{*}{\begin{tabular}[c]{@{}c@{}}Hierarchical\\ Transformer\end{tabular}} &
  \multicolumn{4}{c|}{THUMOS-14~\cite{D1}} &
  \multicolumn{4}{c}{ActivityNet1.3~\cite{D2}} \\ \cline{3-10} 
          &           & 0.3  & 0.5  & \multicolumn{1}{c|}{0.7}  & Avg. & 0.5  & 0.75 & \multicolumn{1}{c|}{0.95} & Avg. \\ \hline
          &           & 60.3 & 45.2 & \multicolumn{1}{c|}{19.9} & 42.6 & 53.0 & 36.0 & \multicolumn{1}{c|}{7.6}  & 35.2 \\
\checkmark &           & 65.5 & 53.4 & \multicolumn{1}{c|}{31.2} & 50.7 & 53.8 & 36.5 & \multicolumn{1}{c|}{7.8}  & 35.8 \\
          & \checkmark & 66.2 & 54.7 & \multicolumn{1}{c|}{34.0} & 52.4 & 53.8 & 36.5 & \multicolumn{1}{c|}{6.9}  & 35.5 \\
\checkmark &
  \checkmark &
  \textbf{71.2} &
  \textbf{61.5} &
  \multicolumn{1}{c|}{\textbf{39.3}} &
  \textbf{58.0} &
  \textbf{53.9} &
  \textbf{36.9} &
  \multicolumn{1}{c|}{\textbf{8.1}} &
  \textbf{36.0} \\ \hline
\end{tabular}%
}
\end{table}

\setlength{\tabcolsep}{7.3pt}
\begin{table}[t]
\centering
\caption{Ablation study on sampling rate in Background Feature Sampling module in terms of mAP(\%).}
\label{tab3}
{\footnotesize
\begin{tabular}{c|clclcc}
\hline
\multirow{2}{*}{Sampling rate $\delta$} & \multicolumn{6}{c}{THUMOS-14~\cite{D1}}                                \\ \cline{2-7} 
                       & 0.3  & 0.4  & 0.5  & 0.6  & \multicolumn{1}{c|}{0.7}  & Avg. \\ \hline
0.3                    & 70.5 & \textbf{67.3} & 59.4 & 50.3 & \multicolumn{1}{c|}{38.0} & 57.1 \\
0.5                    & 70.5 & 66.0 & 59.0 & 49.9 & \multicolumn{1}{c|}{38.7} & 56.8 \\
0.7 & \textbf{71.2} & 67.2 & \textbf{61.5} & \textbf{51.0} & \multicolumn{1}{c|}{\textbf{39.3}} & \textbf{58.0} \\ \hline
\end{tabular}
}
\end{table}

\setlength{\tabcolsep}{5.0pt}
\begin{table}[t]
\centering
\caption{Ablation study results on Transformer design in terms of mAP(\%).}
\label{tab4}
{\footnotesize
\begin{tabular}{c|cccccc}
\hline
\multirow{2}{*}{Type} & \multicolumn{6}{c}{THUMOS-14~\cite{D1}}                                \\ \cline{2-7} 
                      & 0.3  & 0.4  & 0.5  & 0.6  & \multicolumn{1}{c|}{0.7}  & Avg. \\ \hline
CNN                   & 65.5 & 60.4 & 53.4 & 43.1 & \multicolumn{1}{c|}{31.2} & 50.7 \\
Vanilla Transformer  & 68.0 & 63.5 & 55.5 & 45.0 & \multicolumn{1}{c|}{30.2} & 52.4 \\
Hierarchical Transformer & \textbf{71.2} & \textbf{67.2} & \textbf{61.5} & \textbf{51.0} & \multicolumn{1}{c|}{\textbf{39.3}} & \textbf{58.0} \\ \hline
\end{tabular}
}
\end{table}

\section{Experiment}

In this section, we conduct extensive experiments and verify the effectiveness of our method in TAL task. First, we introduce the two primary benchmark datasets: THUMOS14 and ActivityNet 1.3. Then, we show that HTNet achieves state-of-the-art performance in terms of mAP. Finally, we provide additional ablation studies of HTNet.

\subsection{Datasets}
\subsubsection{\textbf{THUMOS14}~\cite{D1}} THUMOS14 consists of 200 and 213 untrimmed videos to validate and test. In addition, it contains 20 action categories for temporal action detection with temporal annotations. The videos contain an average of 15 action instances per video with an average of 8\% overlapping with other instances.

\subsubsection{\textbf{ActivityNet1.3}~\cite{D2}} ActivityNet1.3 consists of 19,994 untrimmed videos. It contains 200 action categories with temporal annotations, and it is split into [training, validation, testing] by the ratio of [2:1:1] following the former setting~\cite{BSN}.

\subsection{Implementation Details}
We extract the video features using two-stream I3D~\cite{I3D}, pretrained on Kinetics~\cite{kinetics}. In THUMOS-14, a snippet contains 16 frames of video with frame stride 4. We set batch size to 2 and trained our model for 45 epochs. In ActivityNet 1.3, a snippet contains 16 frames of video with frame stride 16. We set batch size to 16 and trained our model for 10 epochs. We adopt AdamW optimizer with a learning rate of $10^{-4}$ and use cosine learning rate decay. We use a single TITAN Xp GPU for training and all experiments. 

\subsection{Comparison with the State-of-the-Art Methods}
We present our results on THUMOS-14 and ActivityNet1.3 datasets in Tab.~\ref{tab1}. We report mean average precision (mAP) at different temporal IoU thresholds (tIoU). The tIoU thresholds are \([0.3:0.1:0.7]\) for THUMOS-14 and \([0.5:0.05:0.95]\) for ActivityNet1.3. On THUMOS-14, our method outperforms other state-of-the-art methods with a 58.0\% average mAP, achieving a large improvement of 6.7\% compared with the existing best score at mAP@0.7. On ActivityNet1.3, our method also achieves the competitive result of the highest 36.0\% at average mAP and 36.9\% at mAP@0.75.

\subsection{Qualitative results}
We also provide the qualitative results to demonstrate the effectiveness of our method. We visualize a qualitative comparison of the results from ours and the baseline model AFSD~\cite{AFSD} in Fig.~\ref{qualitative}. The first video explicitly reveals a temporal action boundary; thus, two models successfully localize the action, but ours is slightly close to the ground truth. In the second video, the action in the video is visually not explicit, so the ground truth can confuse the model to localize. The baseline model, which does not consider global context, localizes the more extended boundary than the ground truth. 
In contrast, we can observe that our method successfully detects by sampling the background context. In the last video, it is also tricky to localize specific 'VollyballSpiking' actions because other actions in the video are similar. Nevertheless, compared to the baseline model, our method obtains accurate results, which shows the effectiveness of our method considering the inherent semantic relationships.

\subsection{Ablation Study}

\subsubsection{\textbf{Effectiveness of BFS and Hierarchical Transformer}}
We verify the proposed background feature sampling module and hierarchical transformer by the ablation study in Tab.~\ref{tab2}. The baseline model (the first row) uses a simple convolutional network instead of hierarchical transformer without BFS. On THUMOS-14, the complete model, including the two modules, achieves 58.0\% average mAP, which is increased by 15.4\%, 7.3\%, and 5.6\% over the baseline model, with only BFS, and with only the hierarchical transformer model, respectively.

\subsubsection{\textbf{Effectiveness of sampling rate}}
We conduct an ablation study on the sampling rate of background to verify the effect of different sampling rates \(\delta\). We reported on THUMOS-14 dataset in terms of mAP in Table~\ref{tab3}. The first (\(\delta=0.3\)) and second rows (\(\delta=0.5\)) show low performance, and the last row (\(\delta=0.7\)) shows the best performance. Here, we can find that the performance can be lower if the background region of the sampling is too long.

\subsubsection{\textbf{Effectiveness of Transformer design}}
Further, we study the effects of Transformer design in Tab.~\ref{tab4}. The baseline (without transformer) model uses a simple convolutional network instead of transformer. In the vanilla transformer, we concatenate the multi-scale features as single sequential data and then input it into a single transformer encoder. As a result, the baseline model (using CNN) shows the lowest performance of 41.8\% on average mAP, and the vanilla transformer, which has one encoder, is slightly improved by 0.7\% over the baseline model. Finally, the performance of the proposed hierarchical transformer is improved by 2.0\% over the baseline model. These results represent that the design of adaptation to the multi-scale features contributes to and boosts the performance.

\subsubsection{\textbf{False negative profiling}}
\begin{figure}[t!]
    \centering
    \subfigure[AFSD]{\includegraphics[width=1.0\linewidth]{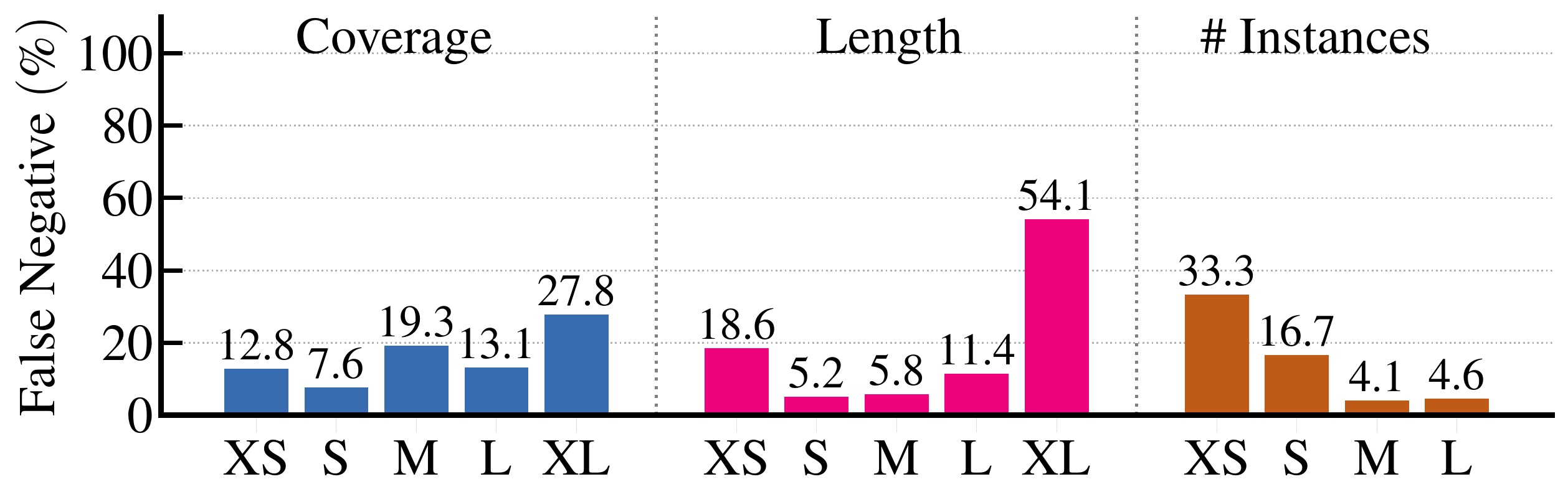}}
    \subfigure[HTNet]{\includegraphics[width=1.0\linewidth]{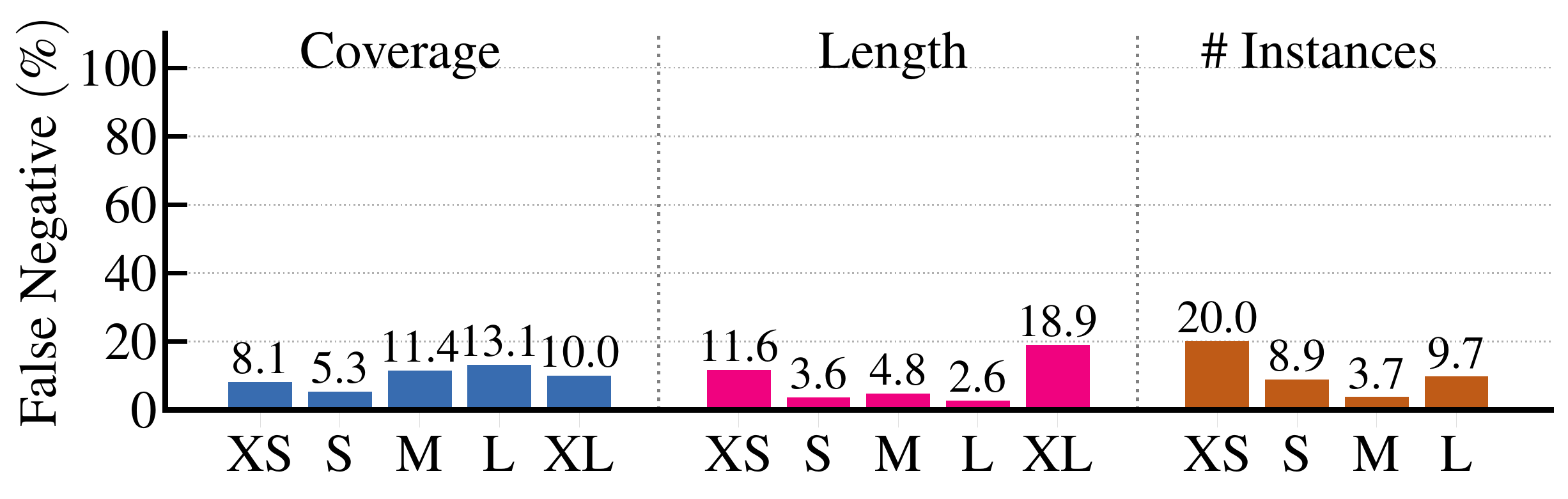}}

    \caption{False negative (FN) profilings of (a) AFSD~\cite{AFSD} and (b) HTNet (ours) results on THUMOS14. Coverage is the relative length of the actions, length is the absolute length of actions, and number of instances denotes the total count of instances.}
    \label{false_negative}
\end{figure}

We provide false negative profiling on THUMOS-14 to verify the effect of our method in Fig.~\ref{false_negative}. We use a diagnosing error tool in temporal action detectors~\cite{FP}. The details of the criteria of categories are presented in~\cite{FP}. In Fig.~\ref{false_negative}, we show that our method successfully mitigates the inability to capture long-range actions while significantly reducing the false negative rates in extra long (XL) action instances compared to AFSD~\cite{AFSD}, where XL denotes action instances longer than 18 seconds.

\section{Conclusion}
In this paper, we presented a novel anchor-free framework based on transformer called HTNet for temporal action localization. HTNet consists of two main modules: (1) The background feature sampling module samples global context information, which enables the refinement of coarse features to include both local boundary information and long-range temporal information. (2) Hierarchical transformer effectively exploits the inherent semantic relationships in a video by self-attention. Our method achieved notable state-of-the-art performance on two benchmark datasets THUMOS-14 and ActivityNet1.3. Although our method performs well, our major limitation is unsatisfactory inference speed in anchor-free methods due to the hierarchical transformer structure. However, we believe simpler and faster transformer structures can be developed in further works.

\bibliography{root}
\bibliographystyle{IEEEtran}

\end{document}